# MODELLING TOURISM DEMAND TO SPAIN WITH MACHINE LEARNING TECHNIQUES. THE IMPACT OF FORECAST HORIZON ON MODEL SELECTION*

*OSCAR CLAVERIA*
*SALVADOR TORRA*
*University of Barcelona (UB)*

*ENRIC MONTE*
*Polytechnic University of Catalunya (UPC)*

This study assesses the influence of the forecast horizon on the forecasting performance of several machine learning techniques. We compare the forecast accuracy of Support Vector Regression (SVR) to Neural Network (NN) models, using a linear model as a benchmark. We focus on international tourism demand to all seventeen regions of Spain. The SVR with a Gaussian radial basis function kernel outperforms the rest of the models for the longest forecast horizons. We also find that machine learning methods improve their forecasting accuracy with respect to linear models as forecast horizons increase. This results shows the suitability of SVR for medium and long term forecasting.

*Keywords:* forecasting, tourism demand, Spain, support vector regression, neural networks, machine learning.

*JEL classification:* C02, C22, C45, C63, E27, R11.

T he increasing weight of the tourism industry in the gross domestic product of most countries regardless of economic fluctuations explains the growing interest in the sector from economic circles. Spain is one of the world's most important destinations after China, France and the United States. The country received close to 65 million tourist arrivals in 2014. The Canary Islands, and were the regions that recorded a greater increase in the number of visitors, with rates above 10%. The steady growth of tourism demand in highlights the importance of correctly anticipating the number of arrivals for the tourism industry.

(*) We would like to thank the Editor, Máximo Camacho, and two anonymous referees for their useful comments and suggestions. This paper has been partially financed by the project ECO2016-75805-R of the Spanish Ministry of Economy and Competitiveness.

In recent years, machine learning (ML) techniques have attracted increasing attention for time series prediction [Hastie *et al.* (2009)]. ML is a field from Artificial Intelligence (AI), and has been central to pattern recognition. ML is based on the design of algorithms to infer structures from a given set of data. ML allows to deal with different problems, such as classification, clustering, and regression. Empirically, it has been proved that rough sets algorithms and fuzzy time series models are particularly indicated for short-term forecasting with limited data [Peng *et al.* (2014)], while Neural Network (NN) and Support Vector Machine (SVM) models are preferable for longer term predictions.

The SVM technique was first developed for classification and pattern recognition (Burges, 1988). Xu *et al.* (2009) use SVMs to improve tourist expenditure classification for visitors to Hong Kong, and Li and Sun (2012) use a SVM-based firm failure prediction model to anticipate the failure of firms in the Chinese tourism industry.

The initial idea has been extended to regression by using the support vectors for local approximation, allowing for non-linear regression estimation in the form of Support Vector Regression (SVR). SVR has been widely used for forecasting purposes in finance [Tay and Cao (2002); Kim (2003); Cao (2003); Pai and Lin (2005); Huang *et al.* (2005); Chen *et al.* (2006)] and other fields [Pai and Hong (2007); Guo *et al.* (2008); Hong (2011) and Zhang *et al.* (2012)].

Nevertheless, very few attempts have been made for tourism demand forecasting [Chen and Wang (2007); Hong *et al.* (2011); Wu *et al.* (2012) and Akin (2015)]. All these studies focus on one-step ahead forecasts at the national level. In this research, we design an experiment to evaluate the forecasting performance of several ML techniques by means of an iterated multi-step ahead forecasting comparison at the regional level.

The main aim of this study is to analyze how the accuracy of SVR and NN predictions is influenced by the forecast horizon under consideration. In order to do so we focus on international tourism demand to all seventeen regions of. We use six different forecast horizons and several SVR and NN models, and a linear model as a baseline. This thorough comparison allows us to shed some light on the suitability of the SVR technique to forecast seasonal time series and tourism demand in particular.

The study proceeds as follows. First, we review the literature on tourism demand forecasting with ML techniques. In the next section the different forecasting methods are described. The data set and the experimental settings are given in the following section. Then, results of the out-of-sample forecasting competition are discussed. Finally, the last section provides a summary of the implications and potential lines for future research.

## 1. LITERATURE REVIEW

A growing body of literature has focused on tourism demand forecasting, but most research efforts apply either casual econometric models [Cortés-Jiménez and Blake (2011); Page *et al.* (2012) and Li *et al.* (2013)] or time series models [Chu (2009); Assaf *et al.* (2011); Gounopoulos *et al.* (2012) and Gunter and Önder (2015)]. See Song and Li (2008), Kim and Schwartz (2013), and Peng *et al.* (2014) for a thorough review of tourism demand forecasting studies. These studies note that the per-formance of the forecasting models varies according to different factors, such as the frequency of the data, the country of origin, the destination, and the length of the forecast horizons. This lack of consensus regarding the most accurate model to forecast tourism demand, has led us to focus the study on data-driven approaches based on ML.

The need for more accurate forecasts has led to an increasing use of ML techniques to obtain more refined predictions of tourist arrivals at the destination level. Goh *et al.* (2008) apply a rough sets approach to forecast U.S. and U.K. tourism demand for Hong Kong. Yu and Schwartz (2006) and Tsaur and Kuo (2011) use fuzzy time series models in predicting tourism demand in Taiwan. Wu *et al.* (2012) note that SVM-based modelling is more indicated to deal with tourism data characteristics. The authors compare the forecasting accuracy of different ML techniques (SVM and Gaussian process regression models) to and ARMA model using monthly tourist arrivals to Hong Kong from thirteen countries of origin from 1985 to 2008, and obtain the most accurate predictions with ML models.

The SVM technique was originally introduced as a classification method following the idea of using a subset of the training samples, known as support vectors to represent the class boundaries in the classification problem (Cristianini and Shawe-Taylor, 2000). The use of these support vectors is related to the solution of an optimization problem that maximizes a margin between classes after a transformation of the data. The optimization problem yields a solution based on the samples aligned along the border between classes.

SVMs are first applied to tourism demand forecasting by Pai and Hong (2005) and Hong (2006), who use a SVM models to forecast tourist arrivals to Barbados, obtaining better forecasting results that with NNs. Velásquez *et al.* (2010) also obtain better forecasts with SVMs than with MLP and ARIMA models for different time series, including monthly totals of international airline passengers.

The original idea behind the SVM mechanism has recently been extended to regression analysis. The introduction of the Vapnik's insensitive loss function, together with the use of genetic algorithms (GAs) for parameter selection, have recently led to increased use of SVRs. Chen and Wang (2007) incorporate a GA in a SVR and compare it to Back Propagation NN and ARIMA models to predict tourist arrivals to China, using quarterly data from 1985 to 2001, and finding evidence in favor of SVRs. Hong *et al.* (2011) compare a SVR with a hybrid chaotic algorithm to forecast annual tourist arrivals to Barbados, and also obtain more accurate forecasts than with ARIMA models.

Akin (2015) compares the forecasting results of SVR to that of SARIMA and NN models to predict international monthly tourist arrivals to Turkey in 2011. By means of a novel approach to model selection based on decision trees, the author finds that SVR outperforms NNs in all cases but SARIMA models only when the slope feature is more prominent.

The most widely used feed-forward NN topology in tourism demand forecasting is the multi-layer perceptron (MLP) [Claveria and Torra (2014); Teixeira and Fernandes (2014); Molinet *et al.* (2015) and Hassani *et al.* (2015)]. A special class of multi-layer feed-forward architecture with two layers of processing is the radial basis function (RBF) network. The first study to implement a RBF NN for forecasting tourism demand is that of Kon and Turner (2005), who use a RBF network model to forecast arrivals to Singapore. Cang (2014) generates predictions of UK inbound tourist arrivals and combines them in non-linear models. Çuhadar, *et al.* (2014) compare the forecasting performance of RBF and MLP NNs. Claveria *et al.* (2015a) find that RBF networks provide better forecasting results than the MLP and Elman architectures. Molinet *et al.* (2015) propose using different periodicities as input variables in NN models for tourism demand forecasting. The authors obtain more precise forecasts for longer horizons than with ARIMA models.

While most of the previous forecasting comparisons focus on one-step ahead predictions at the national level, in this study the forecasting performance of three SVR models is compared to that of two NN models by means of an iterated multi-step ahead forecasting comparison. We assess the forecasting accuracy of the models for different forecast horizons at a regional level. In a recent study, Lehmann and Wohlrabe (2015) address some of the issues related to regional forecasting.

There have been several studies on tourism in Spain at regional level [Aguiló *et al.* (2005); Bardolet and Sheldon (2008); Santana-Jiménez and Hernández (2011); Andrades-Caldito *et al.* (2013) and Sarrión-Gavilán *et al.* (2015)], but only a few regarding tourism demand forecasting. This research mostly focuses on two regions: the Balearic Islands and the Canary Islands.

Regarding tourism demand forecasting to the Balearic Islands, Álvarez-Díaz and Rosselló-Nadal (2010) forecast British tourist arrivals using meteorological variables, Rosselló-Nadal (2001) predicts turning points in arrivals, Garín-Muñoz and Montero-Marín (2007) use a dynamic model with panel data, and Medeiros *et al.* (2008) develop a NN-GARCH model. With respect to the Canary Islands, Hernández-López and Cáceres-Hernández (2007) use a GA with a transition matrix to forecast tourists' characteristics, Gil-Alana *et al.* (2008) models international monthly arrivals using different time-series approaches.

The first attempt to use ML techniques for tourism demand forecasting in Spain is that of Palmer *et al.* (2006), who design a MLP neural network to forecast tourism expenditure in the Balearic Islands. The authors use quarterly data from 1986 to 2000, and find that MLP NN provide more accurate forecasts when data have been detrended and deseasonalized. This result coincides with that of Claveria *et al.* (2017), who analyse the effects of data pre-processing on the forecasting performance of NN models and find that the predictive accuracy of the models improves with seasonal adjusted data. In line with previous research by Pattie and Snyder (1996) and Burger *et al.* (2001), Palmer *et al.* (2006) also find that NNs are especially suitable for long-term forecasting, as long as data is pre-processed.

In a recent study, Claveria *et al.* (2015b) design a multiple-input multiple-output NN framework to generate predictions for all visitor markets to a destination simultaneously. By using monthly data of tourist arrivals to Catalonia from 2001 to 2012, the authors generate forecasts for one, three and six months ahead with three different NN topologies and find that RBF NNs outperform the rest of the models.

## 2. Forecasting models

The main contribution of this study to the previous literature on tourism demand forecasting is the evaluation of the performance of several ML techniques by means of an iterated multi-step ahead forecasting comparison. As most previous research focuses on one-step ahead predictions, in this study we design an experiment to assess how the forecast accuracy of the models is influenced by the forecast horizon under consideration. As ML techniques do not need to pre-process the data, we are able to compare the forecasting performance of three SVR models and two NN topologies using seasonal raw data.

### 2.1. Support Vector Regression

The SVR mechanism proposed by Drucker *et al.* (1997) can be regarded as an extension of SVMs to construct data-driven and non-linear regressions. The original SVM algorithm was developed by Vapnik (1995) and Cortes and Vapnik (1995). The idea behind the technique of SVR is to define an approximation of the regression function within a 'tube' of radius $\varepsilon$ such that its output is as near as possible to the desired output $d_t$:

$$f(x_t) = \omega\varphi(x_t) + b \qquad [1]$$

where $x_i$ is the input vector; $\omega$ is a weight vector; $b$ is a constant.

The parameters of the model are estimated by solving a convex optimization problem that uses as cost function the $\varepsilon-$ insensitive loss function $L_\varepsilon$:

$$L_\varepsilon(d_t, y_t) = \begin{cases} \left| d_t - y_t \right| - \varepsilon, & \left| d_t - y_t \right| \geq \varepsilon \\ 0 & \text{otherwise} \end{cases} \qquad [2]$$

The introduction of two positive slack variables $\xi_t$ and $\xi_t^*$ allows to reformulate the SVR as an optimization problem:

$$Minimize \; \frac{1}{2}\left\|\omega\right\|^2 + C\sum_{t=1}^{n}\left(\xi_t + \xi_t^*\right) \; subjected \; to \begin{cases} d_t - \omega\varphi(x_t) - b \leq \varepsilon + \xi_t \\ \omega\varphi(x_t) + b - d_t \leq \varepsilon + \xi_t^* \end{cases} \xi_t^{(*)}, C \geq 0 \qquad [3]$$

The hyperparameter $\varepsilon$ determines the allowed margin for the regression. The hyperparameter $C$ determines the number of noisy samples that overlap with the tube that contains the regression function, and therefore can be considered as a regularization parameter. As a result, high values of this parameter do not allow for sudden changes in the slope of the regressed function. The selection of the hyperparameters $\varepsilon$ and $C$ is done by means of cross-validation.

To solve [3], we can introduce two Lagrange multipliers and a kernel function $K(x_i, x_t)$ in the decision function. In this study we use three different kernels:

$$\text{– Linear kernel (L-SVR)} \qquad K(x, y) = a_1 x * y + a_2 \qquad [4]$$

$$\text{– Polynomial kernel (P-SVR)} \qquad K(x, y) = \left(a_1 x * y + a_2\right)^h \qquad [5]$$

$$\text{– Gaussian RBF kernel (G-SVR)} \qquad K(x, y) = \exp\left[\left(1/\delta^2\right)\left(x - y\right)^2\right] \qquad [6]$$

Where $a_1$ and $a_2$ are constants, $h$ is the degree of the polynomial kernel, and $\delta^2$ is the bandwidth of the Gaussian RBF kernel. For a comprehensive introduction to SVR see Cristianini and Shawe-Taylor (2000).

Note that the nonlinearities on the kernels do not define the shape of the nonlinearities given by the regression function. The nonlinearities of the kernel function can be regarded as a similarity measure between the sample points and the support vectors (Cristianini and Shawe-Taylor, 2000). Therefore, the role of the kernel is to determine the set of support vectors along the margin that contribute to the final value. The final output consists of a linear combination of the values of the mapping using the kernel function of the test sample with each support vector:

$$y_t = \sum_{i=1}^{N\_sv} \omega_i K\left(x_i, x_t\right) + b \qquad [7]$$

where $N - sv$ is the number of support vectors of the model, $\omega_i$ is the i-th element of weight vector $\omega$, and $K(x_i, x_t)$ is the kernel function that yields a similarity index of the input vector $x_i$ and the i-th support vector.

## 2.2. Neural Networks

Neural networks emulate the processing of human neurological system to identify related spatial and temporal patterns from historical data. A complete summary on the use of NNs with forecasting purposes can be found in Zhang *et al.* (1998). In this study we use two different NN architectures: the RBF and the MLP.

The RBF model can be specified as:

$$y_t = \beta_0 + \sum_{j=1}^{q} \beta_j g_j\left(x_{t-i}\right)$$
$$g_j\left(x_{t-i}\right) = \exp\left(-\sum_{j=1}^{p}\left(x_{t-i} - \mu_j\right)^2 \bigg/ 2\sigma_j^2\right) \qquad [8]$$

Where $\{x_{t-i}; i = 1, ..., p\}$ and $\{\beta_j; \sigma_j; j = 1, ..., q\}$. The output vector is denoted by $y_t$, $x_{t-i}$ is the input value, $g_j$ the activation function, $u_j$ the centroid vectors, $\beta_j$ the weights, and $\sigma_j$ the spread for neuron $j$. We denote $q$ as the number of neurons in the hidden layer, which ranges from 5 to 30, increasing for longer forecast horizons.

The MLP model is given by:

$$y_t = \beta_0 + \sum_{j=1}^{q} \beta_j g\left(\sum_{i=1}^{p} w_{ij}x_{t-i} + w_{0j}\right) \qquad [9]$$

Where $\{x_{t-i}; i = 1, ..., p\}$, $\{w_{ij}; i = 1, ..., p; j = 1, ..., q\}$, $\{\beta_j; j = 1, ..., q\}$. The weights connecting the input with the hidden layer are denoted by $w_{ij}$, while $g$ is the non-linear function of the neurons in the hidden layer. The number of neurons is estimated by cross-validation. A complete summary on the implementation of NNs can be found in Haykin (2008).

3. Data and methodology

Data on international tourist arrivals to at a regional level are provided by the Spanish Statistical Office (National Statistics Institute – INE – *www.ine.es*). Data include the monthly number of foreign tourists arriving to each region (Autonomous Community) over the time period 1999:01 to 2014:03. As the main aim of this study is to assess the forecasting performance of ML techniques for different forecast horizons when using seasonal raw data, we use the level of the series, without any pre-processing on the original data.

In Table 1 we present the number of international tourist arrivals and the frequency distribution of international tourist arrivals in 2013 by region. The main three destinations (Catalonia, the Balearic Islands and Andalusia) account for more than half (59%) of the total number of tourist arrivals to Spain. The first six destinations account for almost 90% of the total number of tourist arrivals, which corroborates that tourism demand is highly concentrated in very few regions.

Table 1: Distribution of the frequency of tourist arrivals to Spanish regions

| Year 2013 | Tourist arrivals | % | % cumulated |
|---|---|---|---|
| Catalonia | 1.0281.308 | 24.95% | 24.95% |
| Balearic Islands | 7.384.863 | 17.92% | 42.87% |
| Andalusia | 6.330.745 | 15.36% | 58.23% |
| Canary Islands | 6.044.595 | 14.67% | 72.90% |
| Madrid (Community) | 4.054.804 | 9.84% | 82.73% |
| Valencia (Community) | 2.701.118 | 6.55% | 89.29% |
| Basque Country | 915.076 | 2.22% | 91.51% |
| Castilla-Leon | 883.526 | 2.14% | 93.65% |
| Galicia | 826.443 | 2.01% | 95.66% |
| Aragon | 400.521 | 0.97% | 96.63% |
| Castilla-La Mancha | 306.395 | 0.74% | 97.37% |
| Navarra | 226.060 | 0.55% | 97.92% |
| Cantabria | 201.297 | 0.49% | 98.41% |
| Murcia (Region) | 196.098 | 0.48% | 98.89% |
| Asturias | 189.320 | 0.46% | 99.35% |
| Extremadura | 181.200 | 0.44% | 99.78% |
| La Rioja | 88.621 | 0.22% | 100.00% |

Source: Own elaboration.

In Figure 1 we map the frequency distribution of international tourist arrivals to Spain by region. The first six destinations are marked in grey. We can see that all these regions, with the exception of Madrid, are located in coastal areas of the Mediterranean, showing the asymmetric concentration of tourism.

Figure 1: MEAN ANNUAL GROWTH RATE OF TOURIST
ARRIVALS TO SPAIN BY REGION (1999-2013)

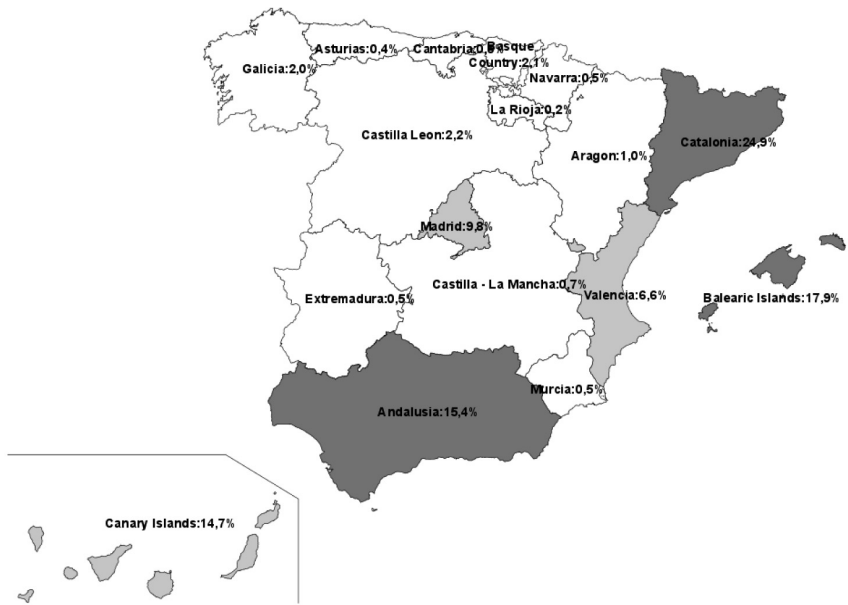

Source: Own elaboration.

Table 2 presents a descriptive analysis of the data. The Balearic Islands and Catalonia are the two regions with the highest peaks. While the Balearic Islands is the region with the highest dispersion in the arrival of tourists, the Canary Islands present the lowest coefficient of variation (CV), also known as relative standard deviation, which is a dimensionless measure of dispersion obtained as the ratio of the standard deviation to the mean and expressed as a percentage. This result may in part be explained by weather conditions, as in the climate is mild and temperatures remain virtually constant throughout the year.

The performance of the different models is assessed by means of an iterated multi-step ahead forecasting comparison. In order to be able to characterize correctly the performance of the ML methods, and to adjust the hyperparameters and the structure of the models, the database is divided into three subsets:

– Training – From 1999:01 to 2006:12
– Validation – From 2007:01 to 2011:12
– Test – From 2012:01 to 2014:01

In applying ARMA models, only two subsets are used: the in-sample (from 1999:01 to 2011:12) and the out-of-sample periods (from 2012:01 to 2014:01). Initially, the first 52% monthly observations are selected as the training set, the next 33% as the validation set, and the last 15% as the test set. Note that the different sets do not overlap in time.

Table 2: DESCRIPTIVE ANALYSIS OF FOREIGN TOURIST ARRIVALS (1999:01-2014:03)

| Region | Minimum | Maximum | Mean | Standard deviation | Coefficient of Variation |
|---|---|---|---|---|---|
| Andalusia | 182.848 | 770.987 | 453.843.7 | 160.241.8 | 35.3% |
| Aragon | 7.901 | 59.194 | 25.868.9 | 11.384.5 | 44.0% |
| Asturias | 2.029 | 33.714 | 11.783.5 | 7.546.5 | 64.0% |
| Balearic Islands | 23.446 | 1.387.491 | 509.102.3 | 423.971.4 | 83.3% |
| Canary Islands | 212.470 | 619.311 | 359.724.3 | 93.466.4 | 26.0% |
| Cantabria | 2.030 | 32.070 | 13.750.8 | 8.552.5 | 62.2% |
| Castilla-Leon | 18.128 | 134.683 | 62.450.4 | 30.444.4 | 48.7% |
| Castilla-La Mancha | 11.483 | 39.308 | 25.856.1 | 8.378.2 | 32.4% |
| Catalonia | 157.103 | 1.442.017 | 625.334.3 | 306.900.6 | 49.1% |
| Valencia (Community) | 80.377 | 322.857 | 171.155.0 | 52.886.8 | 30.9% |
| Extremadura | 4.618 | 31.558 | 12.443.7 | 4.502.2 | 36.2% |
| Galicia | 8.395 | 126.066 | 51.043.9 | 29.595.0 | 58.0% |
| Madrid (Community) | 135.249 | 469.760 | 279.640.7 | 78.578.3 | 28.1% |
| Murcia (Region) | 4.897 | 24.845 | 14.138.4 | 3.999.6 | 28.3% |
| Navarra | 2.592 | 35.152 | 12.748.8 | 7.444.7 | 58.4% |
| Basque Country | 14.388 | 142.644 | 51.169.5 | 25.532.1 | 49.9% |
| La Rioja | 983 | 15.657 | 6.224.5 | 3.534.6 | 56.8% |

Source: Own elaboration.

The validation set is used to adjust the different parameters of the ML models. In the case of NNs, the validation set is used to determine the optimal stopping time for the training, and the topology of the network. In the case of SVR models, the validation set is used for selecting the hyperparameters. The test set is used to estimate the performance of forecasting models on unseen data (Bishop, 1995; Ripley, 1996).

After the first step, the partition between train and test sets is done sequentially: as the prediction advances, forecasts are successively incorporated to the training database, leaving the validation set fixed. Therefore, for each forecast, the size of the test set decreases by one period, while the size of the training set increases by one period. In each iteration, models are retrained, and both the optimal structure and the parameters are obtained from the validation set.

The forecasting accuracy of a SVR model depends in good measure on the value of the hyperparameters. Therefore, given the size of the sample, we do an exhaustive enumeration of all possible combinations to estimate the value of the hyperparameters, using as a selection criterion the performance on the validation set at each step.

Regarding the NN models, once the topology of the models is specified, the estimation of the weights of the networks can be done by means of different algorithms. To avoid the possibility that the search for the optimum value of the parameters finishes in a local minimum, we use a multi-starting technique that initializes the NNs several times for different initial random values, trains the network and chooses the one with the best result on a validation database. All models are implemented using the 'Scikit-learn' Python module, which is an open source library that integrates a wide range of state-of-the-art ML algorithms (Pedregosa *et al.*, 2011).

4. Empirical results

The forecasting performance of the different models is assessed for different time horizons (1, 3, 6 and 12 months) by computing several accuracy measures. First, we obtain the Relative Mean Absolute Percentage Error (rMAPE) statistic for all regions (Table 3). The rMAPE ponders the MAPE of the model under evaluation against the MAPE of the benchmark model. We use an ARMA model as a baseline.

We find that the lowest rMAPE values are obtained for longer forecast horizons, except for NN models (Figure 2). This result indicates that SVR models improve their forecasting performance with respect to ARMA models as forecast horizons increase. With the exception of the Canary Islands and Madrid, which are the only two regions that do not present a seasonal pattern, in most regions NN models outperform linear models for two, three and six-step ahead forecasts, while SVR models for three, six and twelve-step ahead predictions (Figure 2).

Figure 2: Forecast accuracy – Average rMAPE by forecast horizon

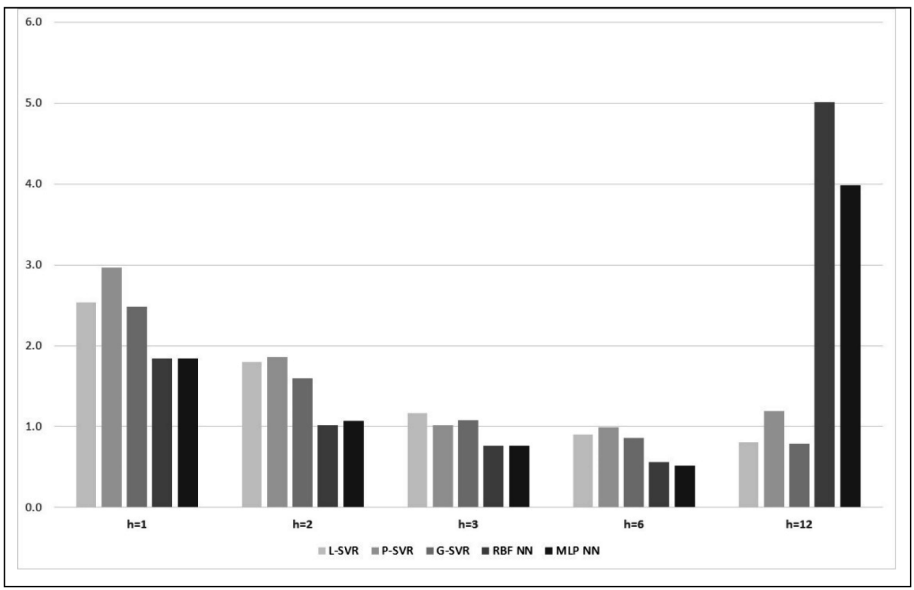

Source: Own elaboration.

In Figure 3 we map the different regions regarding the average rMAPE. We compare the mean rMAPE values of SVR models to those of NN models, regardless of the forecast horizon. The highest values are obtained in the Basque Country, Madrid, the Canary Islands and the Balearic Islands. At the other end, Aragon, Castilla-La Mancha and La Rioja present the lowest average rMAPE values, indicating the regions where, on average, ML models show the best forecasting performance with respect to ARMA models.

Table 3: FORECAST ACCURACY. RELATIVE MAPE
(2013:03-2014:01) – ML VS. ARMA MODELS

| Forecast horizon ($h$) | L-SVR | P-SVR | G-SVR | RBF NN | MLP NN |
|---|---|---|---|---|---|
| Andalusia | | | | | |
| $h = 1$ | 2.202 | 3.019 | 2.182 | 1.468 | 1.507 |
| $h = 2$ | 1.808 | 1.323 | 2.067 | 0.838 | 0.873 |
| $h = 3$ | 1.554 | 1.635 | 1.801 | 0.624 | 0.573 |
| $h = 6$ | 0.898 | 1.175 | 1.011 | 0.506 | 0.441 |
| $h = 12$ | 0.446 | 0.815 | **0.440** | 5.161 | 3.845 |
| Aragon | | | | | |
| $h = 1$ | 2.131 | 1.858 | 1.861 | 1.296 | 1.306 |
| $h = 2$ | 1.398 | 1.445 | 1.188 | 0.775 | 0.852 |
| $h = 3$ | 0.814 | 0.670 | 0.596 | 0.493 | 0.541 |
| $h = 6$ | 0.840 | 0.889 | 0.814 | 0.419 | **0.352** |
| $h = 12$ | 0.891 | 1.010 | 0.876 | 1.790 | 1.598 |
| Asturias | | | | | |
| $h = 1$ | 3.049 | 3.322 | 2.946 | 1.710 | 1.624 |
| $h = 2$ | 1.518 | 1.380 | 1.221 | 0.792 | 0.905 |
| $h = 3$ | 0.618 | 0.428 | 0.458 | 0.441 | 0.474 |
| $h = 6$ | 0.438 | 0.603 | 0.383 | 0.255 | **0.223** |
| $h = 12$ | 0.462 | 0.669 | 0.489 | 4.938 | 3.983 |
| Balearic Islands | | | | | |
| $h = 1$ | 2.559 | 3.801 | 2.888 | 2.149 | 2.000 |
| $h = 2$ | 3.344 | 4.512 | 3.035 | 0.968 | 0.984 |
| $h = 3$ | 1.064 | 0.732 | 1.087 | 0.513 | 0.477 |
| $h = 6$ | 0.199 | 0.479 | 0.187 | 0.214 | **0.098** |
| $h = 12$ | 0.308 | 0.546 | 0.290 | 22.176 | 14.387 |
| Canary Islands | | | | | |
| $h = 1$ | 5.413 | 6.294 | 5.249 | 5.199 | 5.203 |
| $h = 2$ | 3.768 | 3.932 | 3.606 | 3.516 | 3.515 |
| $h = 3$ | 3.303 | 3.449 | 3.264 | 3.285 | 3.359 |
| $h = 6$ | 2.708 | 2.816 | 2.718 | 2.620 | 2.612 |
| $h = 12$ | 1.926 | 2.150 | **1.913** | 5.536 | 5.753 |
| Cantabria | | | | | |
| $h = 1$ | 2.556 | 3.130 | 2.350 | 2.023 | 2.480 |
| $h = 2$ | 1.669 | 2.057 | 1.285 | 0.847 | 0.953 |
| $h = 3$ | 0.822 | 0.457 | 0.370 | 0.510 | 0.508 |
| $h = 6$ | 0.448 | 0.736 | 0.307 | 0.224 | **0.177** |
| $h = 12$ | 0.370 | 0.533 | 0.385 | 3.800 | 2.476 |

Table 3: Forecast accuracy. Relative MAPE
(2013:03-2014:01) – ML vs. ARMA models (continuation)

| Forecast horizon ($h$) | L-SVR | P-SVR | G-SVR | RBF NN | MLP NN |
|---|---|---|---|---|---|
| Castilla-Leon | | | | | |
| $h = 1$ | 2.340 | 2.694 | 2.330 | 1.552 | 1.480 |
| $h = 2$ | 1.692 | 1.864 | 1.539 | 0.866 | 0.934 |
| $h = 3$ | 0.938 | 0.491 | 0.840 | 0.562 | 0.561 |
| $h = 6$ | 0.504 | 0.744 | 0.394 | 0.306 | **0.206** |
| $h = 12$ | 0.361 | 0.684 | 0.353 | 5.259 | 4.473 |
| Castilla-La Mancha | | | | | |
| $h = 1$ | 1.970 | 2.161 | 2.084 | 1.290 | 1.292 |
| $h = 2$ | 1.246 | 1.384 | 1.104 | 0.744 | 0.799 |
| $h = 3$ | 0.966 | 1.402 | 0.833 | 0.601 | 0.599 |
| $h = 6$ | 0.742 | 0.876 | 0.618 | 0.386 | 0.331 |
| $h = 12$ | 0.323 | 0.640 | **0.263** | 1.868 | 1.512 |
| Catalonia | | | | | |
| $h = 1$ | 2.233 | 2.534 | 2.212 | 1.668 | 1.713 |
| $h = 2$ | 1.701 | 1.687 | 1.471 | 0.861 | 0.950 |
| $h = 3$ | 0.970 | 0.685 | 0.729 | 0.551 | 0.552 |
| $h = 6$ | 0.761 | 0.738 | 0.791 | 0.389 | **0.379** |
| $h = 12$ | 0.864 | 0.889 | 0.850 | 5.401 | 4.790 |
| Valencia (Community) | | | | | |
| $h = 1$ | 2.229 | 3.247 | 2.278 | 1.894 | 1.712 |
| $h = 2$ | 1.435 | 1.345 | 1.452 | 1.046 | 1.056 |
| $h = 3$ | 0.922 | 0.801 | 0.922 | 0.759 | 0.763 |
| $h = 6$ | 0.988 | **0.602** | 0.976 | 0.661 | 0.661 |
| $h = 12$ | 0.990 | 1.628 | 0.953 | 3.288 | 2.859 |
| Extremadura | | | | | |
| $h = 1$ | 1.771 | 1.827 | 1.802 | 1.389 | 1.332 |
| $h = 2$ | 1.274 | 1.211 | 1.235 | 0.755 | 0.753 |
| $h = 3$ | 1.002 | 1.214 | 1.408 | 0.621 | 0.566 |
| $h = 6$ | 0.923 | 1.067 | 0.808 | 0.561 | **0.519** |
| $h = 12$ | 0.631 | 0.926 | 0.583 | 2.885 | 2.263 |
| Galicia | | | | | |
| $h = 1$ | 2.714 | 3.114 | 2.634 | 1.675 | 1.678 |
| $h = 2$ | 1.464 | 1.568 | 0.899 | 0.825 | 0.893 |
| $h = 3$ | 0.681 | 0.467 | 0.573 | 0.502 | 0.518 |
| $h = 6$ | 0.506 | 0.657 | 0.400 | 0.295 | **0.233** |
| $h = 12$ | 0.501 | 1.182 | 0.498 | 4.390 | 3.445 |

Table 3: FORECAST ACCURACY. RELATIVE MAPE
(2013:03-2014:01) – ML VS. ARMA MODELS (continuation)

| Forecast horizon ($h$) | L-SVR | P-SVR | G-SVR | RBF NN | MLP NN |
|---|---|---|---|---|---|
| Madrid (Community) | | | | | |
| $h = 1$ | 2.752 | 3.625 | 2.687 | 2.323 | 2.056 |
| $h = 2$ | 1.941 | 2.179 | 1.866 | 1.340 | 1.213 |
| $h = 3$ | 1.900 | 1.175 | 1.922 | 1.225 | 1.155 |
| $h = 6$ | 1.927 | 1.867 | 1.917 | 0.995 | **0.928** |
| $h = 12$ | 2.061 | 3.888 | 2.123 | 4.123 | 3.757 |
| Murcia (Region) | | | | | |
| $h = 1$ | 2.010 | 2.021 | 1.961 | 1.460 | 1.489 |
| $h = 2$ | 1.342 | 1.281 | 1.172 | 0.908 | 0.989 |
| $h = 3$ | 1.080 | 0.953 | 1.042 | **0.708** | 0.746 |
| $h = 6$ | 1.204 | 1.016 | 1.210 | 0.724 | 0.762 |
| $h = 12$ | 1.135 | 1.502 | 1.063 | 1.435 | 1.288 |
| Navarra | | | | | |
| $h = 1$ | 2.251 | 2.573 | 2.178 | 1.590 | 1.582 |
| $h = 2$ | 1.365 | 1.634 | 1.254 | 0.775 | 0.853 |
| $h = 3$ | 0.803 | 0.670 | 0.702 | 0.462 | 0.470 |
| $h = 6$ | 0.697 | 0.912 | 0.564 | 0.303 | **0.245** |
| $h = 12$ | 0.562 | 0.796 | 0.558 | 3.189 | 2.827 |
| Basque Country | | | | | |
| $h = 1$ | 2.189 | 2.250 | 2.130 | 1.551 | 1.776 |
| $h = 2$ | 2.031 | 1.893 | 1.790 | 0.850 | 0.934 |
| $h = 3$ | 1.408 | 1.270 | 1.242 | 0.560 | 0.574 |
| $h = 6$ | 0.863 | 0.730 | 0.944 | 0.467 | **0.466** |
| $h = 12$ | 1.516 | 1.707 | 1.486 | 7.005 | 6.491 |
| La Rioja | | | | | |
| $h = 1$ | 2.679 | 3.024 | 2.409 | 1.161 | 1.166 |
| $h = 2$ | 1.591 | 0.979 | 1.010 | 0.664 | 0.709 |
| $h = 3$ | 0.956 | 0.727 | 0.555 | 0.529 | 0.520 |
| $h = 6$ | 0.740 | 0.973 | 0.515 | 0.276 | **0.227** |
| $h = 12$ | 0.387 | 0.667 | 0.346 | 2.972 | 1.999 |

Note: The rMAPE ponders the MAPE of the model under evaluation against the MAPE of the benchmark model. We use an ARMA model as a baseline. Best model for each region in bold.

Source: Own elaboration.

Figure 3: Forecast accuracy – Average rMAPE
by region (SVR vs. NN models)

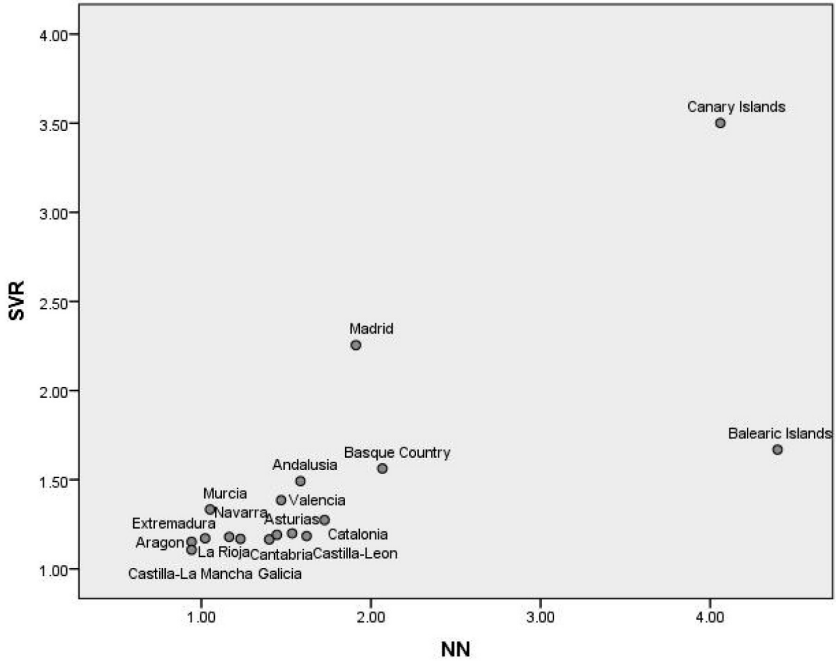

Source: Own elaboration.

To attain a more comprehensive forecasting evaluation, we compute the PLAE statistic proposed by Claveria *et al.* (2015b). The PLAE gives the proportion of out-of-sample periods with lower absolute forecast errors than a benchmark model (Table 4). The PLAE is a dimensionless measure based on the CJ statistic for market efficiency (Cowles and Jones, 1937):

$$PLAE = \frac{\sum_{t=1}^{n} \lambda_t}{n} \quad \text{where} \quad \lambda_t = \begin{cases} 1 & \text{if } |e_{t,A}| < |e_{t,B}| \\ 0 & \text{otherwise} \end{cases} \quad [10]$$

Where $y_t$ is the actual value, and $\hat{y}_t$ the forecasted value at period $t = 1, \dots n$. Forecast errors can then be defined as $e_t = y_t - \hat{y}_t$. Given two competing models *A* and *B*, where *A* refers to the forecasting model under evaluation and *B* stands for the benchmark model. In this study we use the ARMA model as a baseline.

In Table 4 we can observe a similar pattern to the one obtained with the rMAPE. Results show that the G-SVR is the method that outperforms the ARMA model in more cases. Special mention should be made of the Canary Islands and the Community of Madrid, where in most cases no ML method outperforms the ARMA.

Table 4: FORECAST ACCURACY. PLAE
(2013:03-2014:01) – ML VS. ARMA MODELS

| Forecast horizon ($h$) | L-SVR | P-SVR | G-SVR | RBF NN | MLP NN |
|---|---|---|---|---|---|
| Andalusia | | | | | |
| $h = 1$ | 0.182 | 0.091 | 0.091 | 0.091 | 0.091 |
| $h = 2$ | 0.364 | 0.364 | 0.182 | 0.273 | 0.182 |
| $h = 3$ | 0.455 | 0.455 | 0.364 | 0.364 | 0.364 |
| $h = 6$ | 0.545 | 0.364 | 0.364 | 0.273 | 0.455 |
| $h = 12$ | 0.818 | 0.455 | 0.818 | 0.818 | 0.727 |
| Aragon | | | | | |
| $h = 1$ | 0.091 | 0.182 | 0.182 | 0.182 | 0.182 |
| $h = 2$ | 0.273 | 0.273 | 0.273 | 0.455 | 0.273 |
| $h = 3$ | 0.455 | 0.545 | 0.545 | 0.455 | 0.455 |
| $h = 6$ | 0.727 | 0.545 | 0.727 | 0.727 | 0.818 |
| $h = 12$ | 0.636 | 0.364 | 0.636 | 0.636 | 0.818 |
| Asturias | | | | | |
| $h = 1$ | 0.000 | 0.091 | 0.091 | 0.182 | 0.182 |
| $h = 2$ | 0.182 | 0.273 | 0.364 | 0.364 | 0.364 |
| $h = 3$ | 0.545 | 0.636 | 0.727 | 0.455 | 0.636 |
| $h = 6$ | 0.909 | 0.818 | 0.818 | 0.909 | 0.909 |
| $h = 12$ | 0.818 | 0.636 | 0.818 | 0.727 | 0.818 |
| Balearic Islands | | | | | |
| $h = 1$ | 0.182 | 0.182 | 0.364 | 0.273 | 0.273 |
| $h = 2$ | 0.091 | 0.091 | 0.455 | 0.273 | 0.273 |
| $h = 3$ | 0.455 | 0.545 | 0.636 | 0.545 | 0.545 |
| $h = 6$ | 1.000 | 0.909 | 1.000 | 1.000 | 0.818 |
| $h = 12$ | 1.000 | 0.818 | 1.000 | 1.000 | 0.818 |
| Canary Islands | | | | | |
| $h = 1$ | 0.000 | 0.000 | 0.000 | 0.000 | 0.000 |
| $h = 2$ | 0.091 | 0.000 | 0.091 | 0.000 | 0.091 |
| $h = 3$ | 0.000 | 0.000 | 0.000 | 0.000 | 0.000 |
| $h = 6$ | 0.000 | 0.000 | 0.000 | 0.000 | 0.000 |
| $h = 12$ | 0.000 | 0.000 | 0.000 | 0.000 | 0.000 |
| Cantabria | | | | | |
| $h = 1$ | 0.000 | 0.273 | 0.273 | 0.182 | 0.182 |
| $h = 2$ | 0.273 | 0.182 | 0.545 | 0.273 | 0.364 |
| $h = 3$ | 0.727 | 0.818 | 0.727 | 0.636 | 0.636 |
| $h = 6$ | 0.818 | 0.636 | 0.909 | 0.909 | 0.909 |
| $h = 12$ | 0.909 | 0.727 | 0.909 | 0.818 | 0.909 |

Table 4: Forecast accuracy. PLAE
(2013:03-2014:01) – ML vs. ARMA models (continuation)

| Forecast horizon ($h$) | L-SVR | P-SVR | G-SVR | RBF NN | MLP NN |
|---|---|---|---|---|---|
| Castilla-Leon | | | | | |
| $h = 1$ | 0.182 | 0.273 | 0.364 | 0.182 | 0.273 |
| $h = 2$ | 0.364 | 0.182 | 0.364 | 0.455 | 0.364 |
| $h = 3$ | 0.455 | 0.455 | 0.545 | 0.545 | 0.545 |
| $h = 6$ | 0.818 | 0.818 | 0.909 | 0.909 | 0.909 |
| $h = 12$ | 0.909 | 0.727 | 0.909 | 0.909 | 0.818 |
| Castilla-La Mancha | | | | | |
| $h = 1$ | 0.455 | 0.455 | 0.364 | 0.455 | 0.455 |
| $h = 2$ | 0.545 | 0.727 | 0.545 | 0.545 | 0.636 |
| $h = 3$ | 0.545 | 0.455 | 0.818 | 0.818 | 0.636 |
| $h = 6$ | 0.727 | 0.545 | 0.727 | 0.818 | 0.727 |
| $h = 12$ | 0.818 | 0.727 | 0.818 | 0.909 | 0.818 |
| Catalonia | | | | | |
| $h = 1$ | 0.091 | 0.182 | 0.091 | 0.091 | 0.182 |
| $h = 2$ | 0.182 | 0.182 | 0.182 | 0.273 | 0.091 |
| $h = 3$ | 0.364 | 0.545 | 0.455 | 0.364 | 0.364 |
| $h = 6$ | 0.727 | 0.636 | 0.727 | 0.727 | 0.818 |
| $h = 12$ | 0.727 | 0.455 | 0.727 | 0.727 | 0.818 |
| Valencia (Community) | | | | | |
| $h = 1$ | 0.091 | 0.091 | 0.091 | 0.091 | 0.182 |
| $h = 2$ | 0.182 | 0.273 | 0.273 | 0.364 | 0.273 |
| $h = 3$ | 0.545 | 0.545 | 0.545 | 0.545 | 0.545 |
| $h = 6$ | 0.364 | 0.818 | 0.364 | 0.455 | 0.455 |
| $h = 12$ | 0.636 | 0.091 | 0.636 | 0.182 | 0.727 |
| Extremadura | | | | | |
| $h = 1$ | 0.182 | 0.091 | 0.091 | 0.182 | 0.091 |
| $h = 2$ | 0.364 | 0.273 | 0.182 | 0.364 | 0.273 |
| $h = 3$ | 0.545 | 0.545 | 0.182 | 0.455 | 0.727 |
| $h = 6$ | 0.545 | 0.273 | 0.545 | 0.545 | 0.727 |
| $h = 12$ | 0.909 | 0.364 | 0.909 | 0.818 | 0.818 |
| Galicia | | | | | |
| $h = 1$ | 0.091 | 0.091 | 0.091 | 0.182 | 0.182 |
| $h = 2$ | 0.364 | 0.091 | 0.455 | 0.364 | 0.364 |
| $h = 3$ | 0.455 | 0.818 | 0.545 | 0.545 | 0.455 |
| $h = 6$ | 0.818 | 0.909 | 0.818 | 0.909 | 0.909 |
| $h = 12$ | 0.909 | 0.273 | 0.909 | 0.909 | 0.909 |

Table 4: Forecast accuracy. PLAE
(2013:03-2014:01) – ML vs. ARMA models (continuation)

| Forecast horizon ($h$) | L-SVR | P-SVR | G-SVR | RBF NN | MLP NN |
|---|---|---|---|---|---|
| Madrid (Community) | | | | | |
| $h = 1$ | 0.091 | 0.091 | 0.182 | 0.091 | 0.182 |
| $h = 2$ | 0.273 | 0.273 | 0.273 | 0.273 | 0.273 |
| $h = 3$ | 0.182 | 0.545 | 0.182 | 0.182 | 0.182 |
| $h = 6$ | 0.182 | 0.182 | 0.182 | 0.182 | 0.273 |
| $h = 12$ | 0.000 | 0.000 | 0.000 | 0.182 | 0.182 |
| Murcia (Region) | | | | | |
| $h = 1$ | 0.091 | 0.182 | 0.091 | 0.182 | 0.364 |
| $h = 2$ | 0.273 | 0.182 | 0.455 | 0.273 | 0.455 |
| $h = 3$ | 0.273 | 0.455 | 0.364 | 0.364 | 0.364 |
| $h = 6$ | 0.364 | 0.364 | 0.364 | 0.182 | 0.364 |
| $h = 12$ | 0.455 | 0.182 | 0.636 | 0.545 | 0.636 |
| Navarra | | | | | |
| $h = 1$ | 0.091 | 0.000 | 0.091 | 0.182 | 0.091 |
| $h = 2$ | 0.273 | 0.182 | 0.273 | 0.273 | 0.273 |
| $h = 3$ | 0.455 | 0.545 | 0.636 | 0.455 | 0.455 |
| $h = 6$ | 0.727 | 0.455 | 0.727 | 0.636 | 0.818 |
| $h = 12$ | 0.727 | 0.455 | 0.727 | 0.636 | 0.818 |
| Basque Country | | | | | |
| $h = 1$ | 0.182 | 0.091 | 0.182 | 0.182 | 0.000 |
| $h = 2$ | 0.182 | 0.273 | 0.182 | 0.273 | 0.182 |
| $h = 3$ | 0.455 | 0.364 | 0.364 | 0.364 | 0.455 |
| $h = 6$ | 0.545 | 0.636 | 0.545 | 0.545 | 0.636 |
| $h = 12$ | 0.091 | 0.091 | 0.091 | 0.273 | 0.091 |
| La Rioja | | | | | |
| $h = 1$ | 0.273 | 0.000 | 0.182 | 0.182 | 0.182 |
| $h = 2$ | 0.273 | 0.364 | 0.455 | 0.182 | 0.182 |
| $h = 3$ | 0.545 | 0.727 | 0.545 | 0.636 | 0.636 |
| $h = 6$ | 0.636 | 0.455 | 0.818 | 0.818 | 0.818 |
| $h = 12$ | 0.727 | 0.545 | 0.818 | 0.636 | 0.909 |

Note: The PLAE ratio measures the proportion of out-of-sample periods with lower absolute errors than the baseline model (ARMA model). Values below 0.5 indicate that the baseline model displays a higher number of lower absolute forecast errors than the model under evaluation for the out-of-sample period.

Source: Own elaboration.

To test whether the reduction in MAPE is statistically significant between the best three models, in Table 5 we present the results of the Diebold-Mariano (DM) statistic of predictive accuracy (Diebold and Mariano, 1995). The null hypothesis of the test is that the difference between the two competing series is non-significant. A negative sign of the statistic implies that the second model has bigger forecasting errors.

Table 5: DM LOSS-DIFFERENTIAL TEST STATISTIC FOR PREDICTIVE ACCURACY

| 12 months ahead forecasts | G-SVR vs. P-SVR | G-SVR vs. MLP NN |
|---|---|---|
| Andalusia | **-4.004** | -1.728 |
| Aragon | -1.590 | 0.819 |
| Asturias | **-2.596** | -0.083 |
| Balearic Islands | -1.408 | -1.728 |
| Canary Islands | **-2.318** | **2.365** |
| Cantabria | -1.604 | -0.476 |
| Castilla-Leon | **-4.432** | -1.157 |
| Castilla-La Mancha | -1.692 | -2.022 |
| Catalonia | -1.164 | 0.588 |
| Valencia (Community) | **-5.674** | 0.944 |
| Extremadura | -3.136 | -0.388 |
| Galicia | **-6.461** | 0.503 |
| Madrid (Community) | **-11.628** | -0.834 |
| Murcia (Region) | **-2.184** | 0.672 |
| Navarra | **-3.682** | 0.051 |
| Basque Country | **-2.903** | -0.278 |
| La Rioja | **-2.490** | -1.063 |

Note: Diebold-Mariano test statistic with NW estimator. Null hypothesis: the difference between the two competing series is non-significant. A negative sign of the statistic implies that the second model (P-SVR, MLP NN) has bigger forecasting errors. The 5% level critical value is 2.028. Significant values in bold.

Source: Own elaboration.

When comparing the forecast accuracy of the G-SVR against that of the other two models that yielded the lowest MAPE values, we find that the G-SVR shows a significant improvement over the P-SVR in most regions, but the improvements over the MLP NN are not significant.

These results confirm previous research by Hong (2006) and Chen and Wang (2007), who obtain better forecasting results with SVM-based models than with NNs for tourist arrivals to Barbados and China respectively. Velásquez *et al.* (2010) also obtain better predictions with SVMs than with MLP NNs. Notwithstanding, we find that the improvements in terms of MAPE of SVR models with respect to MLP networks are not statistically significant.

Finally in Table 6 we present a summary of the MAPE values. The results show that while linear models are preferable for short-term forecasting, ML techniques are more suitable for long-term prediction. From three-month ahead on, both SVR and NN models outperform ARMA models. In this experiment all ML techniques have been specified so as to use only one temporal lag for concatenation, as opposed to linear models which were set to select the optimal number of AR and MA terms by means of the Akaike information criterion (AIC). The implementation of a model criterion to identify the best suited specification regarding the number of lags in ML methods would allow to perform equivalent forecasting comparisons to linear models.

Table 6: FORECAST ACCURACY − SUMMARY OF MAPE VALUES BY TECHNIQUE

| Forecast horizon ($h$) | Mean | Median |
|---|---|---|
| Linear models (ARMA models) | | |
| $h = 1$ | **0.325** | **0.200** |
| $h = 2$ | 0.472 | 0.315 |
| $h = 3$ | 0.713 | 0.425 |
| $h = 6$ | 0.727 | 0.457 |
| $h = 12$ | 0.661 | 0.349 |
| Support Vector Regression (SVR models) | | |
| $h = 1$ | 0.903 | 0.464 |
| $h = 2$ | 0.974 | 0.473 |
| $h = 3$ | 0.619 | 0.401 |
| $h = 6$ | 0.416 | 0.360 |
| $h = 12$ | **0.402** | **0.318** |
| Neural Networks (RBF and MLP) | | |
| $h = 1$ | 0.655 | 0.420 |
| $h = 2$ | 0.751 | 0.428 |
| $h = 3$ | 0.707 | 0.413 |
| $h = 6$ | 0.424 | 0.342 |
| $h = 12$ | **0.389** | **0.299** |

Note: Lowest values for each type of technique in bold.

Source: Own elaboration.

To summarize, the overall forecasting performance of all methods improves for longer forecast horizons. Regarding the different techniques we obtain slightly better predictions with SVR models than with NNs. Nevertheless, not all SVRs show the same performance. The SVR with a Gaussian RBF kernel outperforms the rest of the models, especially for mid and long-term forecasts. This result is indicative that the data is clustered along the margins that surround the regression function in the SVR model. The fact that the data is clustered instead of spread in the input space, also explains the lower performance of the polynomial kernel.

As Gaussian RBF kernels are the most prevalent choice for SVR-based forecasting (Smola and Schölkopf, 1998), and are also easy to implement, our results suggest the potential of this SVR model for tourism demand forecasting. These findings also highlight the importance of properly selecting the kind of kernel function when using SVR models with forecasting purposes.

5. SUMMARY AND CONCLUSIONS

As more accurate predictions become essential for effective policy planning, new forecasting methods provide room for improvement. Artificial intelligence methods based on machine learning such as Support Vector Regressions and Artificial Neural Networks have attracted increasing interest to refine the predictions in the tourism industry. SVR and NN are flexible techniques that admit a wide array of models. The main purpose of this study is to assess the forecasting performance of Support Vector Regression for tourism demand prediction and seasonal forecasting. With this aim we assess the influence on the prediction accuracy of the different techniques as forecast horizons increase.

Three different SVR models based on different kernel functions and two alternative NN architectures have been used to generate predictions of international tourism demand to all seventeen regions of Spain. By means of several forecast accuracy measures, the different ML techniques have been evaluated with respect to a linear model used as a baseline. The forecasting out-of-sample comparison shows that the SVR with a Gaussian RBF kernel outperforms the rest of the models in most cases. The statistically significant improvement in terms of forecast accuracy with respect to other SVR models illustrates the importance of not overlooking the parameter and kernel function selection for SVR modeling.

Another interesting finding is that ML methods improve their forecasting performance with respect to linear stochastic models as forecast horizons increase. While NN models show the best performance for intermediate forecast horizons, SVR models tend to yield the best predictions for the longer time horizons. This result suggests that the optimal forecasting methodology depends on the time horizon of the predictions. In this sense future research should focus on developing a model selection criterion dependent on different forecast horizons.

This research contributes to the economic literature and to the tourism industry by shedding some light on the most appropriate SVR model to predict seasonal time series and tourism demand. A question that requires to be examined in further detail is the implementation of model selection criteria for ML techniques that would allow to refine AI-based predictions. The comparison of SVR to alternative ML techniques such as Gaussian process regression is also left for future research.

RESUMEN

El presente estudio evalúa la influencia de los horizontes predictivos sobre la precisión de las predicciones obtenidas mediante técnicas de inteligencia artificial basadas en aprendizaje automático. Para ello se compara la capacidad predictiva de diferentes modelos de Regresión de Soporte Vectorial (RSV) y de Redes Neuronales (RNA) con un modelo lineal utilizado como referencia. El análisis se centra en la demanda de turismo extranjero en España a nivel regional. La RSV entrenada con un kernel de función de base radial gaussiana supera al resto de modelos para las predicciones a más largo plazo. También se encuentra que los métodos de aprendizaje automático mejoran su capacidad predictiva con respecto a los modelos lineales a medida que aumenta el horizonte de predicción. Este resultado pone de manifiesto la idoneidad de la RSV para la predicción a medio y largo plazo.

*Palabras clave:* predicción, demanda turística, España, regresión de soporte vectorial, redes neuronales artificiales, técnicas de aprendizaje automático.

*Clasificación JEL:* C02, C22, C45, C63, E27, R11.